\documentclass[letterpaper, 10 pt, conference]{ieeeconf}  

\IEEEoverridecommandlockouts 

\usepackage{graphics} 
\usepackage{epsfig} 
\usepackage{mathptmx} 
\usepackage{times} 
\usepackage{amsmath} 
\usepackage{amssymb}  
\usepackage{subcaption}
\usepackage{algpseudocode}
\usepackage{algorithmicx}
\usepackage{algorithm}
\usepackage{graphicx}
\usepackage{multicol}
\usepackage{multirow}
\usepackage[table]{xcolor}
\newcommand{\rev}[1]{#1}
\usepackage{booktabs}
\usepackage{censor}
\usepackage[utf8]{inputenc}
\title{\LARGE \bf
\rev{Memory-Aware Keyframe Management for Dense Maps\\on Memory-Constrained Robots} }

\author{Huichang Yun, Jaeyeul Kim, and Seungho Yoo$^{*}$%
 \thanks{Dept. of Computer Engineering, Pukyong National University, Busan, Korea
 {\small hcyun@pukyong.ac.kr, \{jykim, shyoo\}@pknu.ac.kr}}%
 \thanks{$^{*}$ Corresponding Author}%
}

\begin{document}

\maketitle
\thispagestyle{empty}
\pagestyle{empty}

\begin{abstract}
\rev{Running a language model and a dense 3D map on one device means the two share one memory, and exceeding its limit is an out-of-memory failure.
Map systems that keep a bounded working set bound it by a count of keyframes, but the memory one keyframe occupies varies widely with what it carries, so a count does not control what the working set occupies.
The unit of admission is as poorly matched, since a radius around the pose, a fixed cell, or the graph neighbors of the last matched keyframe all move parts of the map that are not needed.
We therefore propose two decisions.
First, the memory the resident map uses is bounded directly, and the cost of every admission is known before any data is read, from per-keyframe resident sizes computed offline and carried with the map. Second, keyframes are admitted and released at a semantic spatial unit, the rooms and corridors of the building, so that the robot holds the zone it occupies together with the zone it is about to enter while the zones it has passed are released.
Every admission measures the resident total, makes room, and only then reads, so the budget is never overshot.
On two NVIDIA Jetson boards the budget holds in every run and the model gives up at most 0.8 percent of its throughput, whereas without it the operating system kills the model or the map.
Under that budget, and over the environments evaluated, admission by semantic zone localizes 6--9 points more of the query frames than a geometric partition of matched zone count and about twice as much as the retrieval of RTAB-Map, at a lower ATE.
Bounding memory lets the map and the model coexist, and moving the map in semantic zones makes the bounded map usable.}
\end{abstract}



\section{INTRODUCTION}
\rev{With recent advances in robotics and large-scale Artificial Intelligence (AI) models, mobile robots can provide more detailed services that are closely related to daily life.
To deliver such services within a given area a robot needs persistent spatial memory and accurate localization~\cite{Pangaliman2026,Anwar2024ReMEmbR,Yang2025ThreeDMem}, and high-capacity 3D dense maps are commonly used to supply both.
Built on that foundation, powerful models and rich indoor context let a robot interpret language instructions that refer to places and objects (e.g., ``Go to the table next to the sofa in the hallway'') and execute them against a consistent 3D map frame.}

\rev{However, deploying AI models together with large 3D map resources on robots is challenging because they operate in more resource-constrained environments than the workstations or servers where such models are commonly used, leading to strict memory limits.
Systems that keep only part of the map resident bound it by a count of keyframes, but one keyframe costs anything from tens of kilobytes to several megabytes, so a count does not bound the memory the device actually measures and the boundary cannot be placed exactly.
Under tight memory constraints, retrieval at the granularity of a single keyframe is repeated whenever the robot meets a new area, which raises the frequency of load and unload operations and the overhead they carry.
More importantly, on a device where the model and the map share one memory, an out-of-memory failure can occur once the available memory is insufficient.}

\begin{figure}[t]
\centering
\begin{subfigure}{0.80\linewidth}
    \includegraphics[width=\linewidth]{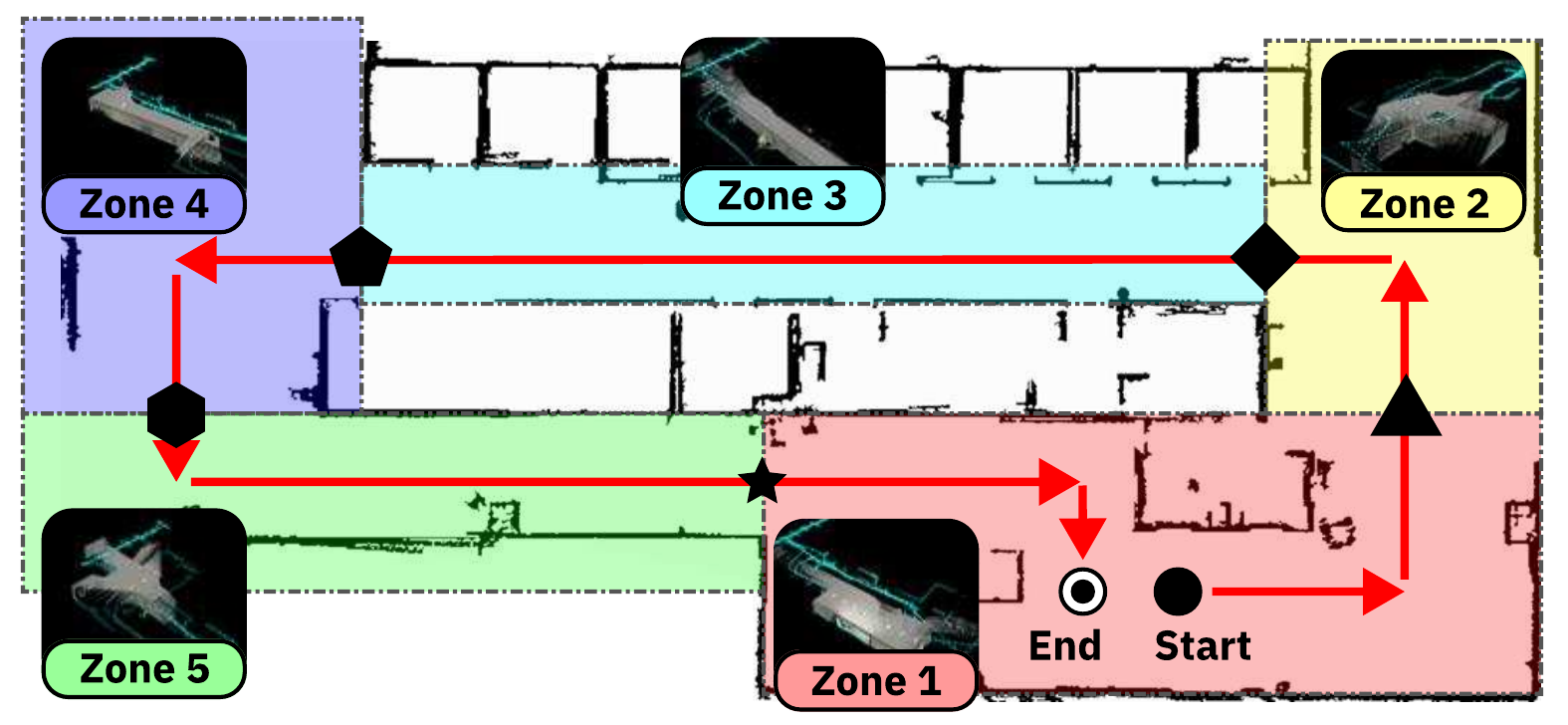}
    \caption{\rev{Semantic zones}}
    \label{fig:fig1a}
\end{subfigure}

\vspace{4pt}
\begin{subfigure}{0.80\linewidth}
    \includegraphics[width=\linewidth]{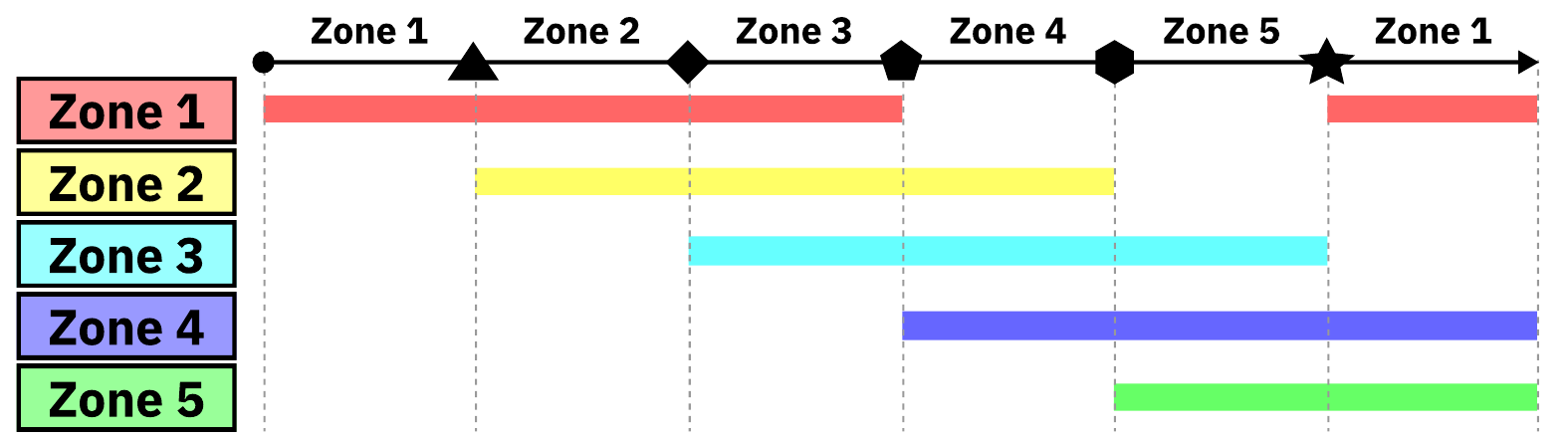}
    \caption{\rev{Managed zones for the robot's progress}}
    \label{fig:fig1b}
\end{subfigure}
\caption{\rev{Zone-granular residency under a memory budget, the concept of ZEBRA}}
\label{fig:system_overview}
\end{figure}

\rev{To mitigate the issues discussed above, we propose ZEBRA, a semantic zone-based map management approach that enables stable 3D dense map utilization under strict memory constraints.
The idea is to treat the map not one keyframe at a time but in semantically meaningful units, the rooms and corridors of the building.
The map is given the memory the language model leaves free, and the resident total is kept under it at every cycle, so the two run together without an out-of-memory failure.
Within that memory the robot holds the zone it currently occupies and the zones it is about to traverse, while the zones it has passed are released.
The resident map therefore stays inside the memory it was given while still covering the space the robot actually observes.
Figure~\ref{fig:system_overview} illustrates the proposed mechanism.
Figure~\ref{fig:fig1a} shows the environment partitioned into semantic zones together with the robot's planned route, and Figure~\ref{fig:fig1b} shows how only a subset of zones is activated as the robot progresses.}

\rev{To realize and validate our proposed approach in large-scale indoor environments, we built on a simulation environment with Nvidia Isaac simulation, and implemented our proposed methods on two Jetson boards, the AGX Orin with 32\,GB of unified memory and the Orin Nano with 8\,GB~\cite{jetson_agx_orin,jetson_orin_nano}, with a model of the Gemma or Qwen family answering queries throughout every run, at the versions listed in Table~\ref{tab:devices}, alongside RTAB-Map~\cite{Labbe2019}.
Across every device and map pairing the budget held, the model gave up at most 0.8 percent of the throughput it reaches with no map payload resident, and under the same budget admitting and releasing keyframes by semantic zone localized more of the mission than the keyframe-level retrieval the system already has.}

The contributions of this paper are as follows.

\begin{itemize}
    \item \rev{A design that bounds the map by the memory the device actually
    has. Existing systems bound the resident map by a count of keyframes, which
    does not control what it occupies. We enforce the limit in memory itself and
    obtain stable operation with no out-of-memory failure and at most
    0.8 percent of the throughput given up by the model beside the map.}
    \item \rev{Semantic zones as the unit of admission and release. A fixed
    grid, and the retrieval RTAB-Map already has, move parts of the map that are
    not needed. The boundaries of a semantic zone roughly follow boundaries of
    visibility, so the same memory holds the space the robot actually sees, and
    more of the mission is localized.}
    \item \rev{An evaluation that separates the two contributions. On the
    AGX Orin and the Orin Nano, over two maps and several budgets, we evaluate
    the bound by comparing bounded against unbounded operation, and the unit of
    admission by comparing semantic zones, cubic chunks and the keyframe-level
    retrieval of the underlying system under one budget. What the second
    comparison shows is that the semantic partitions localize more of the query
    frames than a geometric partition of matched zone count in these
    environments, and that the size of a zone, the number of keyframes it holds,
    has to be weighed alongside the unit.}
\end{itemize}

\section{RELATED WORK}
A dense map that does not fit can be made smaller by discarding parts of it, made lighter by replacing its contents with a representation from which they are reconstructed, or left intact and only partially resident, with the remainder on disk and retrieved when needed.
The three are often combined.
We review them in turn and state, for each, which of our two claims it leaves open.

\subsection{Reducing or Compressing the Map}
\rev{The oldest response to map growth is to keep less of the map, and pose graphs are pruned by information-theoretic criteria \cite{Kretzschmar2011} or by preserving algebraic connectivity \cite{Doherty2024MAC}, redundant poses are marginalized \cite{Yu2024SLIM}, keyframes are selected under a budget \cite{Thorne2025}, and commercial systems keep one pose per grid cell \cite{Noh2024CLOi}.
S-Graphs~2.0 \cite{Bavle2025} uses the room as the unit, marginalizing every keyframe of a room but the first once its walls are detected, so the unit is ours and the operation is not, since the marginalization is permanent and what it saves is optimization time.
A second group keeps the information but stores it differently, using Gaussian mixtures \cite{Li2024GMMap} or wavelet coefficients \cite{Reijgwart2023Wavemap} in place of occupancy grids, multiresolution submaps for neural implicit maps \cite{Tian2025MISO}, and quantized, codebook-stored Gaussian primitives \cite{Lee2024Compact3DGS}.}

\rev{Neither group addresses either claim of this paper.
Pruning changes how many keyframes exist and compression changes what one of them costs, but neither states a bound the device can enforce, neither decides which keyframes should be resident at a given moment, and what pruning discards does not come back when a service above the map later asks for that fidelity.
Compression is therefore orthogonal to residency rather than opposed to it.}

\subsection{Bounding What Is Resident}
\label{subsec:rw_resident}
\rev{The third group leaves the map intact and limits how much of it occupies memory at once.
RTAB-Map, the system we build on, keeps a working memory beside a database-backed long-term memory and transfers nodes between them to bound the time of loop closure detection, so the working memory bounds the candidate space and a keyframe leaves it with its features \cite{Labbe2019,Labbe2013}.
Edge-SLAM keeps a local map of fixed size on the robot and delegates the global map to a server \cite{BenAli2020EdgeSLAM}, and DB-LIO releases the point clouds outside an active window and reloads them on return \cite{Kim2026DBLIO}.
DiskChunGS divides a scene into ten-meter chunks, holds in video memory only those visible from the current keyframe, and evicts the rest to disk under a least-recently-used policy \cite{Feldmann2026}.
A subset of this group operates, as we do, against a map built beforehand and never extended at run time.
B\"urki et al. select which landmarks to query by appearance, with summarization holding the map to a fixed number of landmarks \cite{Burki2018MapManagement}, and Scucchia and Maltoni predict the region the robot is in to pre-select the nodes place recognition will match against, on the same RTAB-Map memory management we extend \cite{Scucchia2024}.
Both decide what to consult rather than what to hold, and both count what they bound.}

\rev{Two properties separate this group from what we claim.
What they bound is a count, a window, or a set of chunks chosen by visibility, and none of these binds memory except through an exchange rate that belongs to the map and to what its keyframes carry rather than to the system that sets the bound, while the resident set also holds keyframes that no admission asked for.
A recent survey reaches the same premise, noting that the size of a map on disk predicts poorly what a system needs at runtime \cite{Pangaliman2026}.
The unit of admission is a radius or a fixed cell, so none of these systems has to decide how the environment should be divided, which makes such a unit inexpensive to adopt, adequate outdoors and inadequate inside a building.
The properties we ask of a partition in Section~\ref{subsec:properties} follow from these two observations.}

\rev{DiskChunGS~\cite{Feldmann2026} is the closest prior system and not one we can compare against directly, since it pages the Gaussians it renders inside an online reconstruction and reports rendering quality, whereas we page the keyframes that localization consumes against a map built beforehand and report whether a pose was produced at all.
The three differences that remain are the ones this paper is about, a budget in bytes rather than counts, a unit of admission that is a room or corridor rather than a ten-meter cube, and a read that admits the next zone on the route before the robot reaches its boundary, where localization would otherwise be interrupted (Section~\ref{subsec:admission}).}

\subsection{What Consumes the Dense Map}
\rev{A growing line of work builds spatial context from the point cloud itself rather than from pose alone, and consumes that geometry directly.
ConceptGraphs represents every object by a 3D point cloud with a semantic feature \cite{Gu2024ConceptGraphs}, HOV-SG derives its floor and room nodes from the point cloud of the whole environment \cite{Werby2024HOVSG}, and Hydra extracts a metric-semantic mesh from which its places and rooms are built \cite{Hughes2022Hydra}.
The bound they state is partial, since Hydra windows only its volumetric layer while the mesh and places it extracts accumulate as a persistent representation \cite{Hughes2022Hydra}.
This is the layer the payload is kept for and the layer this paper is meant to serve, since a keyframe reduced to sparse features is enough to produce a pose and not enough to answer a question about the space.
What such a system gains here is the bound on residency it does not state, and what it supplies in return is the partition, a composition we return to in the conclusion.}

\begin{figure*}[t]
    \centering
    \includegraphics[width=0.72\textwidth]{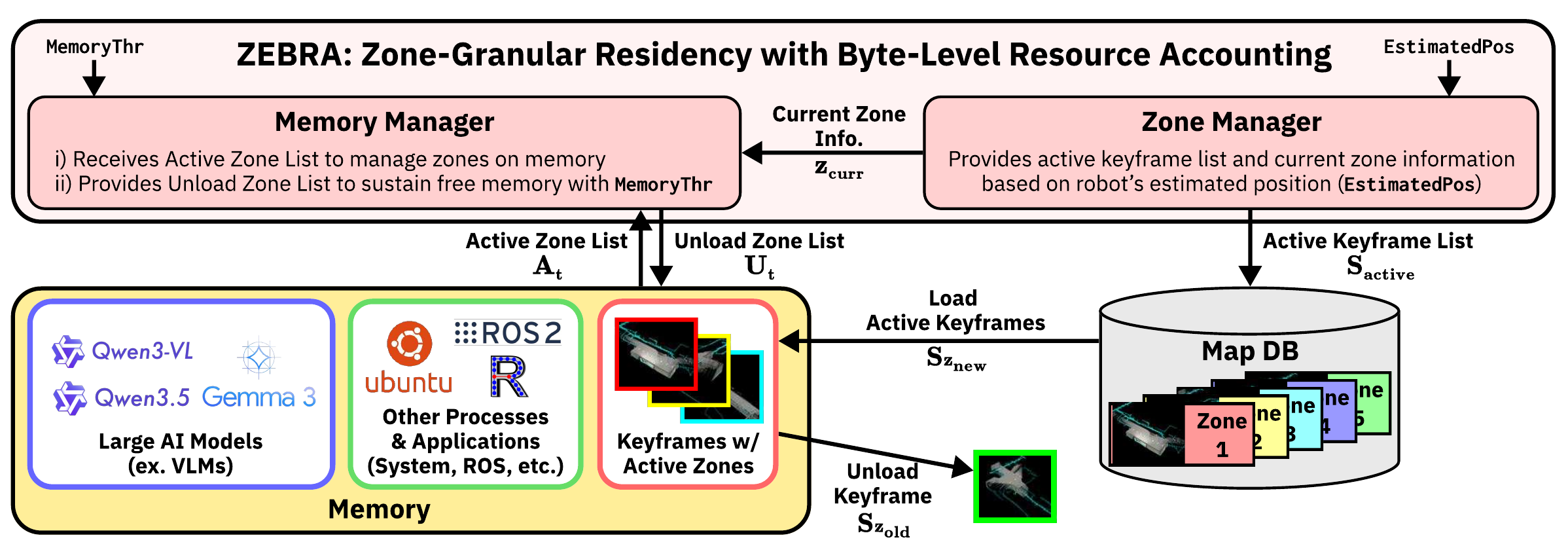}
    \caption{\rev{Overview of ZEBRA}}
    \label{fig:keyframe_management_overview}
\end{figure*}

\section{ZEBRA: ZONE-GRANULAR RESIDENCY UNDER A MEMORY BUDGET}
A robot that localizes against a pre-built dense map cannot hold the entire map in memory, so keyframes must move between the database and the working set while the robot operates.
This section answers two questions, which keyframes should be resident at each instant, and how the resident set is kept inside a budget the platform shares with a running AI model.

Two properties of the deployment make this harder than a cache replacement problem.
First, the quantity that must be bounded is memory and not a count, since a keyframe carrying a decoded point cloud costs roughly five megabytes and one reduced to sparse features about sixty kilobytes, so a budget counted in keyframes means a different amount of memory in every system that sets one and again when that system changes what a keyframe carries.
Second, a wrong decision costs more than a cache miss, because a keyframe absent when the robot needs it cannot be fetched within the same cycle and the pose is lost until the robot reaches a region that is still resident.

We address both by making the unit of admission a semantic zone and the unit of accounting a byte.
\rev{We call the resulting map-management layer ZEBRA, for Zone-Granular Residency with Byte-Level Resource Accounting.}

\subsection{Zones, Payload, and the Resident Set}
\label{subsec:zones}
We assume the environment is partitioned into a finite set of zones $Z$, each corresponding to a spatially meaningful region such as a room or a corridor.
Each keyframe $s_i$ belongs to one zone by its location, written $z(s_i) \in Z$.
The keyframes of a zone form the set $S_z = \{s_i \mid z(s_i) = z\}$.

Let $b(s_i)$ denote the resident size of keyframe $s_i$ in bytes, the compressed image and depth data together with the point cloud decoded at decimation $d$.
Because the decoded payload is generated on demand and never stored in the database, $b(s_i)$ cannot be read at runtime, so we compute it offline and carry it with the map as metadata indexed by keyframe identifier, and the cost of admitting a zone is known exactly before any data is read.
The resident size of a zone is $B_z = \sum_{s_i \in S_z} b(s_i)$.

The map $z \mapsto S_z$ reaches the runtime as an offline table, and how it was defined is invisible to the procedure.
\rev{A pose is mapped to the zone it falls in.}
\rev{Knowledge of where the robot is going enters in one place, as a planned route, the ordered list of zones the mission will traverse, the return leg included.
The runtime keeps an index into this list and advances it to the first of the next four entries that matches the current position, so that a zone passed twice on an out-and-back mission resolves correctly.
The route serves two decisions only, both stated in Section~\ref{subsec:admission}, which zone to read before the robot reaches it and which zones not to retire, and the pose-to-zone map and the sets $S_z$ are the same with or without it.}

Let $W_t$ denote the working set at time $t$, that is, the keyframes actually resident in memory, and let
\begin{equation}
B(W_t) = \sum_{s_i \in W_t} b(s_i)
\label{eq:resident_bytes}
\end{equation}
denote its size in bytes.
The budget $B_{\max}$ is a fixed number of bytes supplied at start-up.
The objective of the runtime procedure is to keep $B(W_t) \leq B_{\max}$ at the end of every cycle while retaining the keyframes the robot is about to need.

We measure $B(W_t)$ over the working set itself and not over the activated zones, because short-term memory, the neighbors pulled in during loop-closure verification and virtual nodes are resident without belonging to any zone.

\subsection{What a Partition Must Satisfy}
\label{subsec:properties}
We require two properties of the partition.
The procedure below does not depend on the algorithm that produced it.

\noindent\emph{Fit.} Every zone must be small enough that admitting it does not by itself exceed the budget.
A partition whose largest zone exceeds $B_{\max}$ cannot be admitted whole, and we handle that case as an exception, admitting the part of the zone that fits (Section~\ref{subsec:admission}).

\noindent\emph{Visibility alignment.} Zone boundaries should approximately coincide with boundaries of visibility, so that the keyframes left outside the active zones are also the ones the robot cannot currently observe, which rooms and corridors give because a wall terminates both the region and the line of sight.

\rev{A partition into cubic chunks, cells of one fixed side laid over the map,} satisfies fit.
With a suitable \rev{side} it approximately satisfies visibility alignment, but only where the environment happens to be divided at that scale.
That limit appears in the evaluation, where \rev{chunks} matched to a semantic partition on zone count \rev{cut} across rooms and corridors alike.

\subsection{Zone-Granular Admission Under a Memory Budget}
\label{subsec:admission}
\rev{Figure~\ref{fig:keyframe_management_overview} shows the runtime structure.
The Zone Manager maps the current pose to the zone it falls in.
The Memory Manager admits and retires zones so that the resident bytes stay within the budget, the two exchange the lists of active and retired zones $A_t$ and $U_t$, and the memory the budget applies to is the memory the map shares with the model and the system processes.}

\rev{Admission proceeds in three steps, and their order is what keeps the resident total inside the budget.
Before any data is read the free room $B_{\max} - B(W_t)$ is computed from the metadata, and if it is smaller than the requested zone, resident keyframes that belong to no active zone and are not immune are released first and zones are retired only if that is not enough.
Only then are keyframes read, because reading first and correcting afterward would allow a brief overshoot, and on a platform that shares memory with an inference server a brief overshoot is an allocation failure.}

\rev{Retirement follows recency of use, so the zone that has gone longest without the robot occupying it goes first, and recency is refreshed whenever the pose maps to a zone, so a zone in use does not age.
When a route is given the victim search skips the zone the robot stands in and the zones still ahead, which changes the order of retirement and not the amount of room it can recover.
Two constraints bound what may be released, the first being the immune set}
\begin{equation}
I_t = S_{z_{\mathrm{curr}}} \cup \rev{(S_{z_{\mathrm{next}}} \cap W_t)} \cup \mathrm{STM}_t \cup H_t \cup P_t,
\label{eq:immune}
\end{equation}
where $S_{z_{\mathrm{curr}}}$ holds the keyframes of the robot's zone, $S_{z_{\mathrm{next}}} \cap W_t$ those keyframes of the next zone on the route that have already been read ahead, $\mathrm{STM}_t$ short-term memory, $H_t$ the nodes of the loop closure hypothesis under verification, and $P_t$ the nodes on the path ahead.
\rev{Each is held because releasing it costs more than it frees, since a verification loses its correspondence, a path becomes invalid, or a keyframe read ahead is read again next cycle.
Retirement also stops while at least one zone remains, and when even then the free room is smaller than the requested zone, which is the case when that zone alone exceeds $B_{\max}$, the procedure admits the part of it that fits, in identifier order, the only case in which zone granularity is broken.}

\rev{With a route, the zone that follows the current one is read before the robot reaches it, under two conditions checked every cycle.
The current zone must be fully resident, so that a prefetch never lengthens the read the robot needs now, and the current and the next zone must fit the budget together, $B_{z_{\mathrm{curr}}} + B_{z_{\mathrm{next}}} \leq B_{\max}$, failing which the next zone waits for its boundary.
When both hold, the read is spread over the traversal, each cycle appending at most $K$ keyframes of the next zone after those of the current zone, with $K = 8$, so that when room is short they are the first to be dropped.}

Algorithm~\ref{alg:zone_admission} states the procedure, whose invariant is that $B(W_t) \leq B_{\max}$ holds at the end of every cycle, so the memory the map occupies is held strictly inside the budget.

\begin{algorithm}[t]
\caption{Zone admission under a memory budget}
\label{alg:zone_admission}
\begin{algorithmic}[1]
\State \textbf{Input:} pose $\hat{\mathbf{x}}_t$ (\texttt{EstimatedPos}), active zones $A_t$,
       working set $W_t$, budget $B_{\max}$ (\texttt{MemoryThr}),
       \rev{next zone on the route $z_{\mathrm{next}}$}
\State $z_{\mathrm{curr}} \leftarrow \textsc{ZoneAt}(\hat{\mathbf{x}}_t)$
\State $A_t \leftarrow A_t \cup \{z_{\mathrm{curr}}\}$; \textsc{Touch}($A_t$, $z_{\mathrm{curr}}$)
\State \rev{$S_{z_{\mathrm{new}}} \leftarrow S_{z_{\mathrm{curr}}} \setminus W_t$}
\If{\rev{$S_{z_{\mathrm{new}}} = \emptyset$ \textbf{and} $B_{z_{\mathrm{curr}}} + B_{z_{\mathrm{next}}} \leq B_{\max}$}}
    \State \rev{$S_{z_{\mathrm{new}}} \leftarrow \textsc{First}_K(S_{z_{\mathrm{next}}} \setminus W_t)$}
\EndIf
\State $I_t \leftarrow \textsc{Immune}(z_{\mathrm{curr}}, z_{\mathrm{next}})$ \Comment{Eq.~\ref{eq:immune}}
\State $r \leftarrow B_{\max} - B(W_t)$
\State \rev{\textsc{Release}($W_t \setminus (I_t \cup \bigcup_{a \in A_t} S_a)$) \textbf{until} $r \geq B(S_{z_{\mathrm{new}}})$, re-measuring $r$ after each release}
\While{\rev{$r < B(S_{z_{\mathrm{new}}})$ \textbf{and} $A_t \setminus \{z_{\mathrm{curr}}\} \neq \emptyset$}}
    \State \rev{$z_{\mathrm{old}} \leftarrow \textsc{LRU}(A_t)$}
    \State \rev{$A_t \leftarrow A_t \setminus \{z_{\mathrm{old}}\}$; $U_t \leftarrow U_t \cup \{z_{\mathrm{old}}\}$}
    \State \textsc{Release}($S_{z_{\mathrm{old}}} \setminus I_t$)
    \State $r \leftarrow B_{\max} - B(W_t)$
\EndWhile
\State \textsc{Load}(\textsc{FitPrefix}($S_{z_{\mathrm{new}}}$, $r$))
\end{algorithmic}
\end{algorithm}

\subsection{Implementation}
We implement the procedure inside RTAB-Map, which provides the three memories and the interfaces for releasing keyframes and reloading them on revisit.
Pose estimation, loop-closure detection and graph optimization are unchanged, and only the decision of which keyframes are resident is replaced.

The behavior is set by the budget $B_{\max}$, the file holding the offline size metadata $b(\cdot)$, and whether the decoded point cloud is attached to a resident keyframe and at what decimation.
\rev{The route, the prefetch switch, and the prefetch batch size $K$ travel in the same file as the zone table, so a plain run and a route-conditioned run differ in that file alone.}

\begin{figure}[t]
\centering
\begin{subfigure}{0.93\linewidth}
    \includegraphics[width=\linewidth]{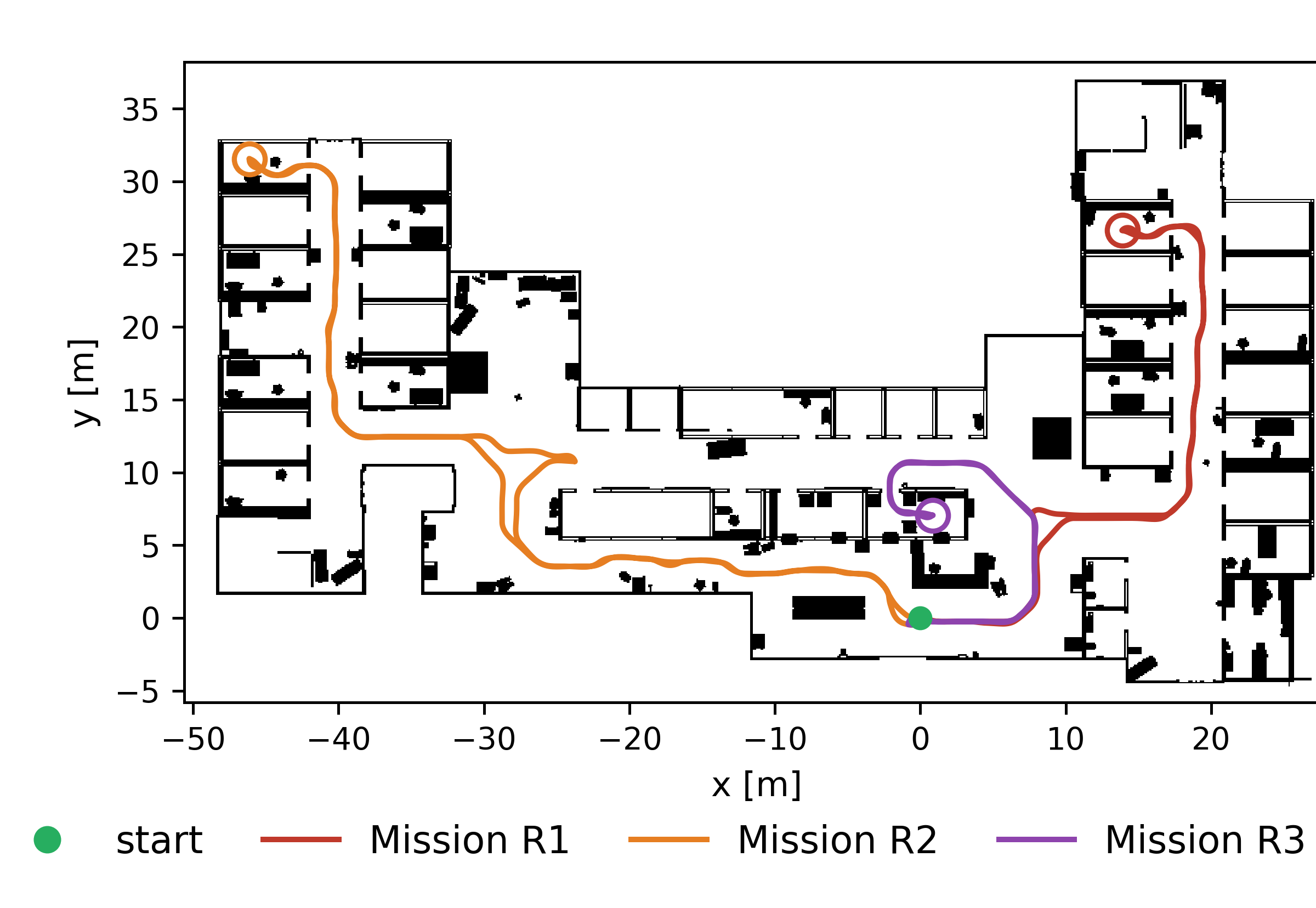}
    \caption{\rev{Hospital}}\label{fig:route_hosp}
\end{subfigure}

\vspace{3pt}
\begin{subfigure}{0.93\linewidth}
    \includegraphics[width=\linewidth]{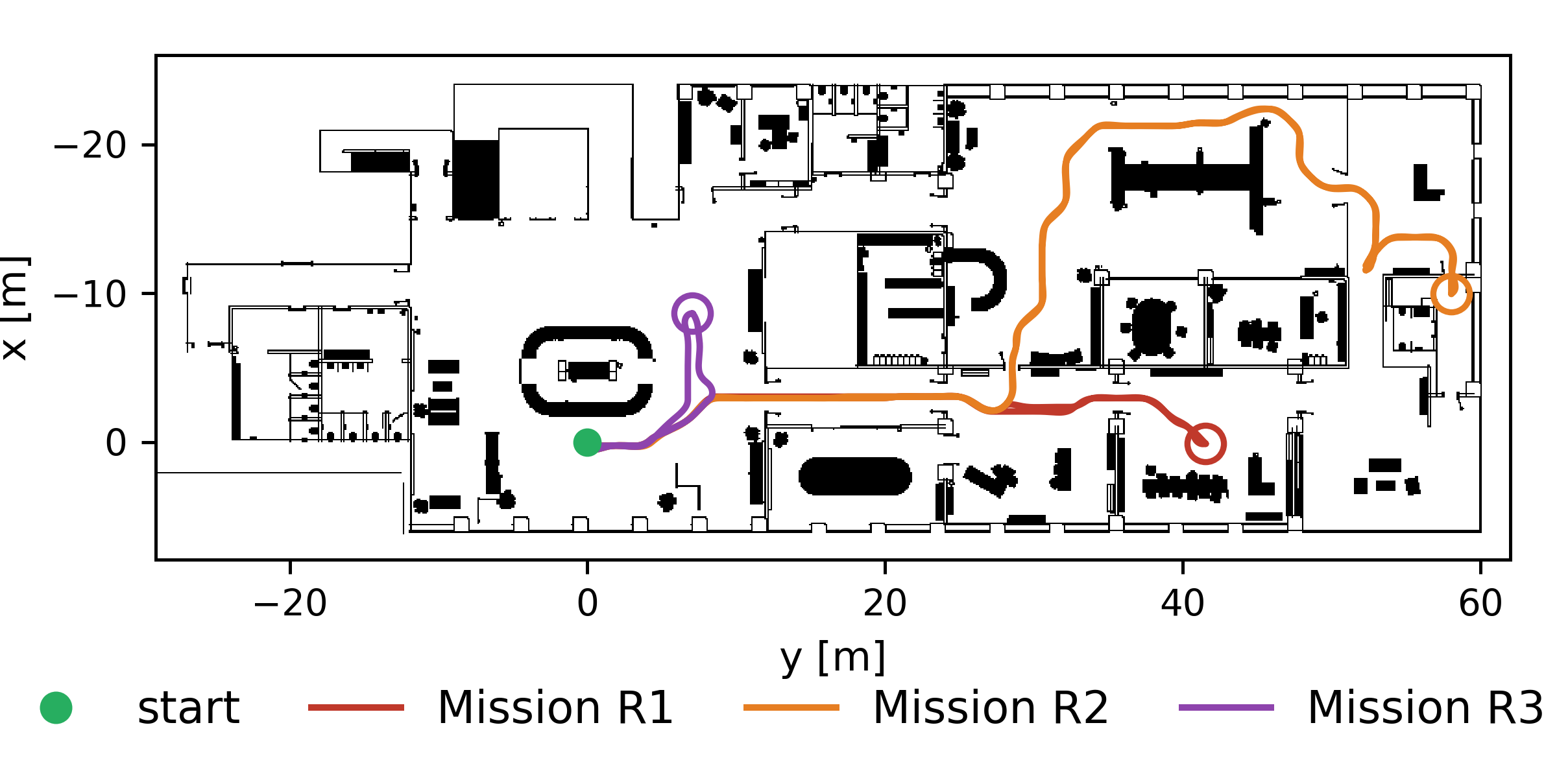}
    \caption{\rev{Office}}\label{fig:route_office}
\end{subfigure}
\caption{\rev{Mission routes R1--R3 on the two maps.}}
\label{fig:routes}
\end{figure}

\begin{figure}[t]
\centering
\begin{subfigure}{0.465\linewidth}
    \includegraphics[width=\linewidth]{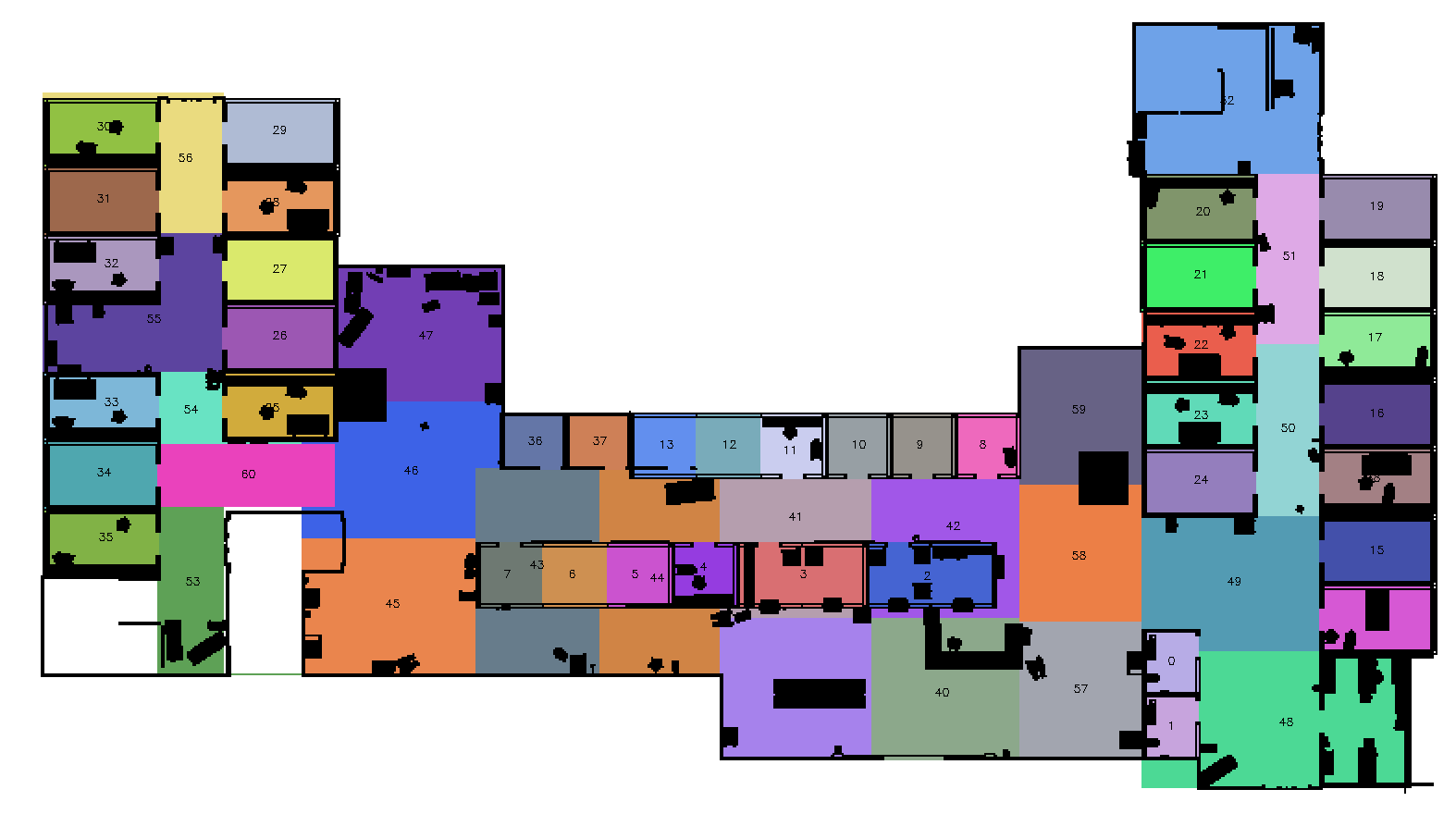}
    \caption{\rev{reference, hosp., 61 zones}}\label{fig:part_ref}
\end{subfigure}\hfill
\begin{subfigure}{0.465\linewidth}
    \includegraphics[width=\linewidth]{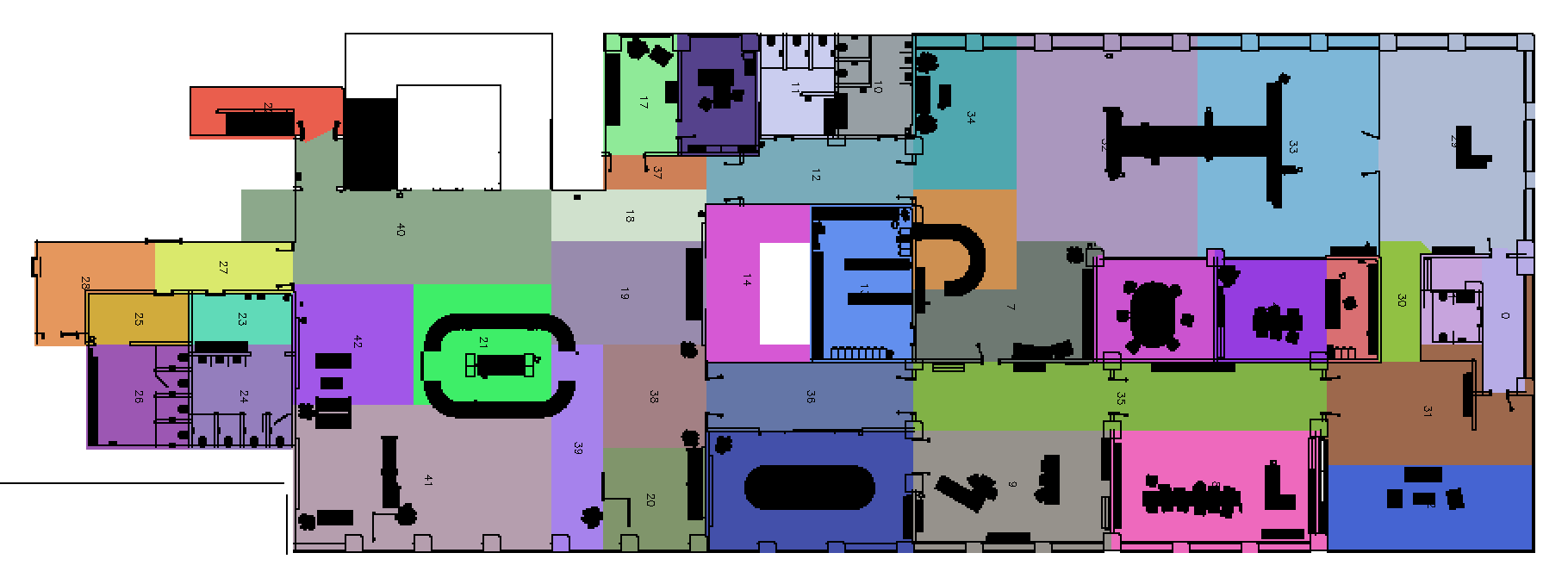}
    \caption{\rev{reference, off., 43 zones}}
\end{subfigure}

\vspace{1pt}
\begin{subfigure}{0.465\linewidth}
    \includegraphics[width=\linewidth]{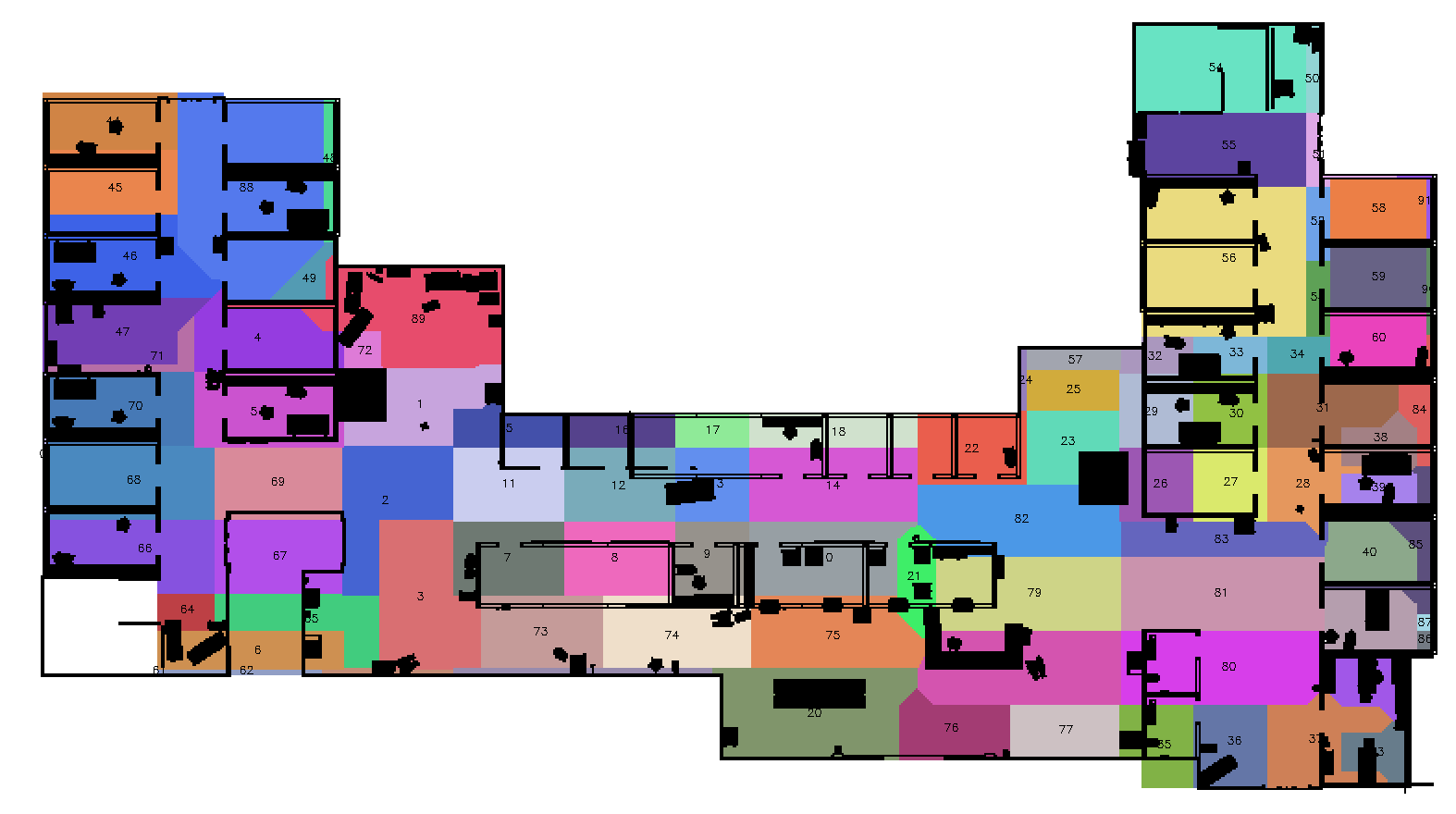}
    \caption{\rev{VLM, hosp., 92 zones}}\label{fig:part_vlm}
\end{subfigure}\hfill
\begin{subfigure}{0.465\linewidth}
    \includegraphics[width=\linewidth]{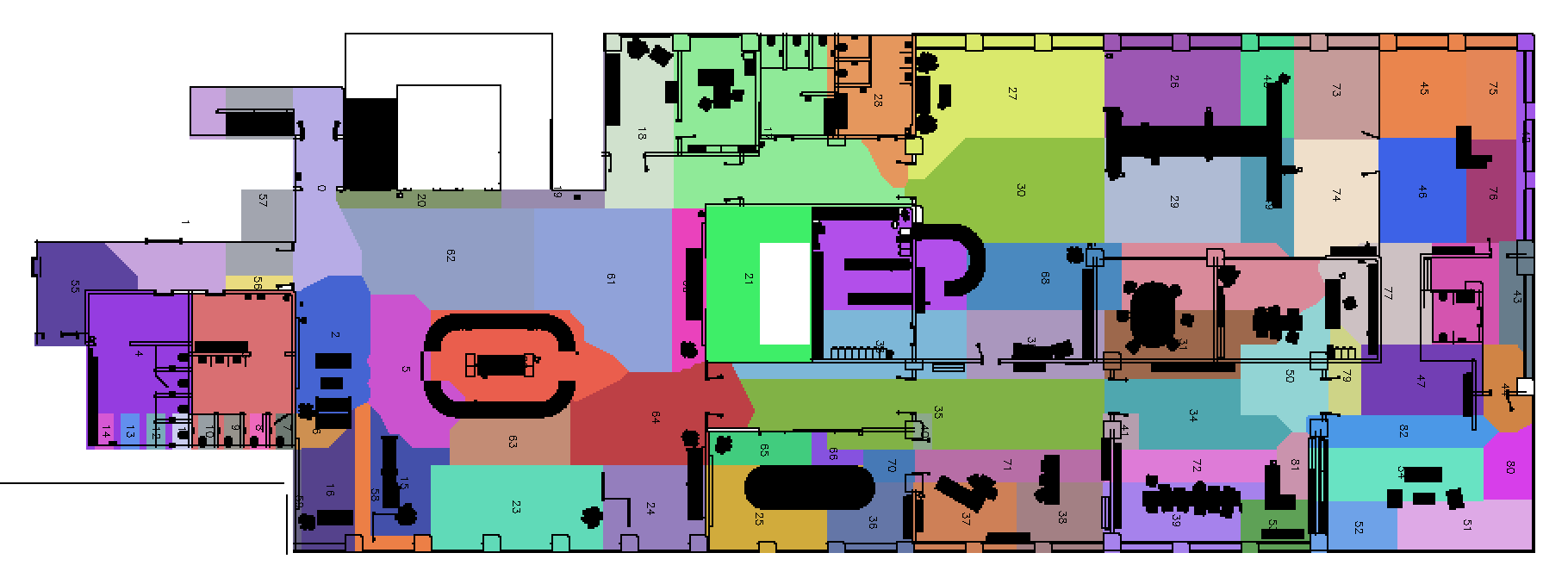}
    \caption{\rev{VLM, off., 83 zones}}
\end{subfigure}

\vspace{1pt}
\begin{subfigure}{0.465\linewidth}
    \includegraphics[width=\linewidth]{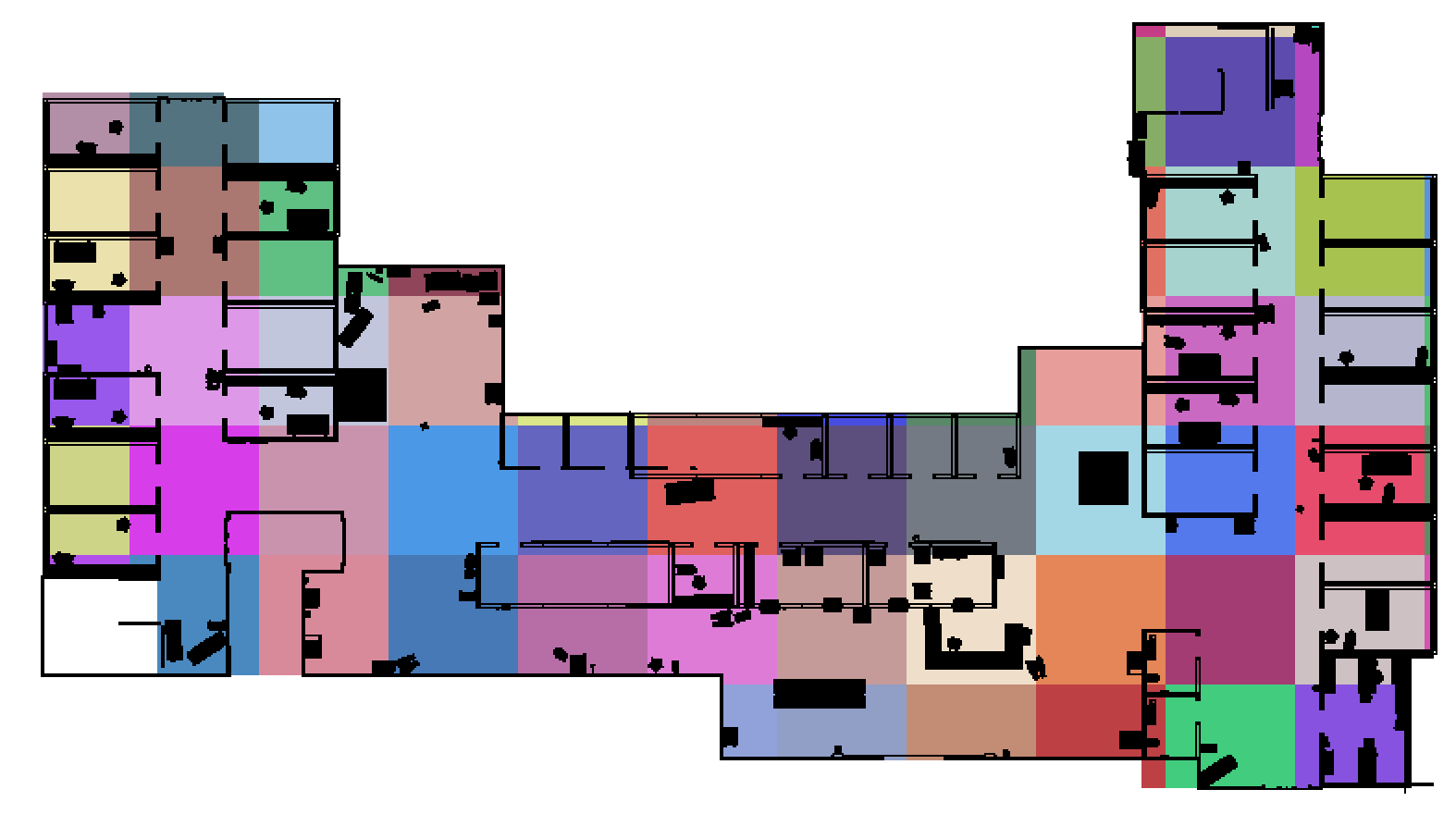}
    \caption{\rev{chunk, hosp., 7\,m, 59 zones}}\label{fig:part_uniform}
\end{subfigure}\hfill
\begin{subfigure}{0.465\linewidth}
    \includegraphics[width=\linewidth]{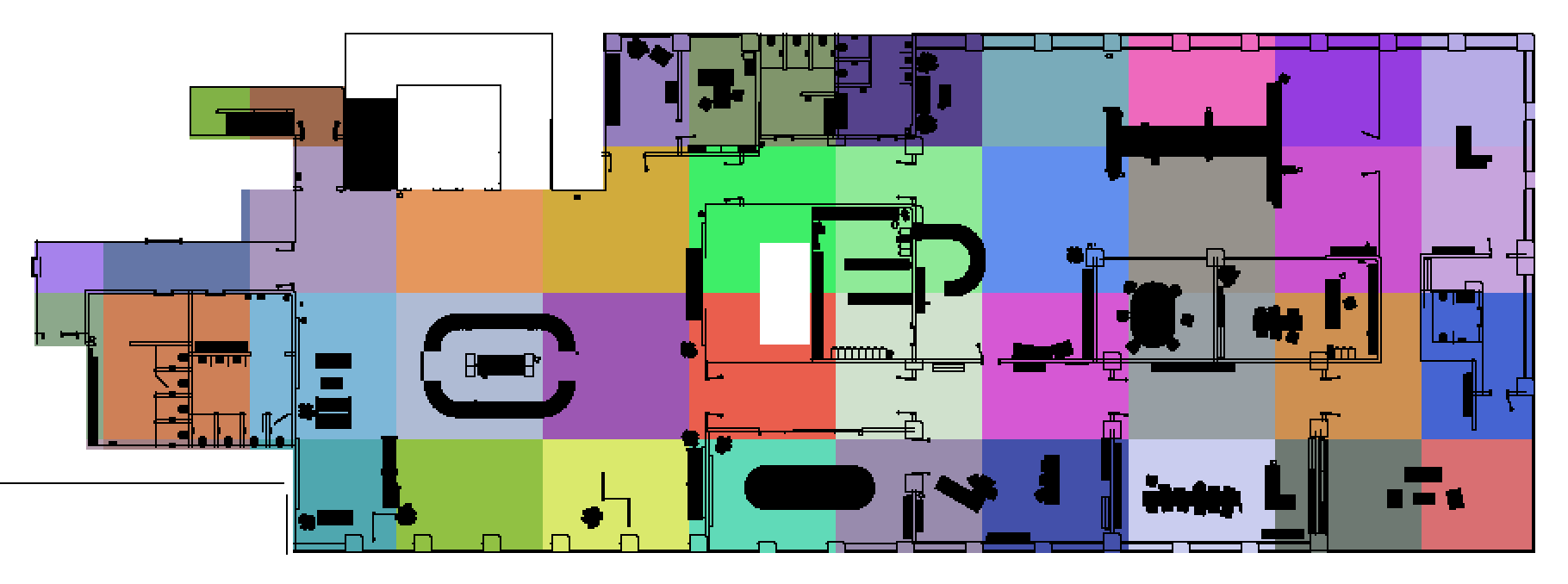}
    \caption{\rev{chunk, off., 8.5\,m, 42 zones}}
\end{subfigure}
\caption{\rev{The three partitions, hospital left and office right.}}
\label{fig:partitions}
\end{figure}

\begin{table}[t]
\centering
\caption{Devices, models, maps, and the residual budget given to the map, the memory the model leaves free. Model size is the resident size of the model. $^{*}$Text-only.}
\label{tab:devices}
\scriptsize
\setlength{\tabcolsep}{1.6pt}
\begin{tabular}{@{}llrlrrr@{}}
\toprule
Device & Model & Model size & Map & Keyframes & Payload & Residual \\
       &       & (GB) &       & (\#) & (MB) & (MB) \\
\midrule
AGX Orin, 32\,GB & Qwen3.5 27B     & 25.0 & Hospital & 1620 & 8399 & 1750 \\
AGX Orin, 32\,GB & Qwen3.5 27B     & 25.0 & Office   & 1041 & 5263 & 2100 \\
Orin Nano, 8\,GB & Gemma3 1B$^{*}$ &  2.6 & Hospital & 1620 & 8399 & 2400 \\
Orin Nano, 8\,GB & Gemma4 E2B      &  4.2 & Office   & 1041 & 5263 &  500 \\
\bottomrule
\end{tabular}
\end{table}

\begin{table*}[t]
\centering
\caption{\rev{Localization rate with respect to the unit of retrieval, mean over the three routes, percent of query frames with a pose.}}
\label{tab:mission}
\footnotesize
\setlength{\tabcolsep}{5pt}
\renewcommand{\arraystretch}{1.15}
\begin{tabular}{@{}lrrrrrrrrrrrrr@{}}
\toprule
& \multicolumn{3}{c}{AGX Orin, hospital} & \multicolumn{3}{c}{AGX Orin, office} & \multicolumn{3}{c}{Orin Nano, hospital} & \multicolumn{3}{c}{Orin Nano, office} & \\
\cmidrule(lr){2-4}\cmidrule(lr){5-7}\cmidrule(lr){8-10}\cmidrule(lr){11-13}
Map budget (MB) & 1700 & 1650 & 1600 & 2050 & 2000 & 1950 & 2350 & 2300 & 2250 & 450 & 400 & 350 & Overall$\uparrow$ \\
\midrule
Org: RTAB-Map         & \underline{45.1} & 31.1 & 34.3 & 5.9 & 7.3 & 7.7 & 39.1 & 30.5 & 15.0 & 6.8 & 8.5 & 13.9 & 20.4 \\
Org: RTAB-Map, route  & 34.7 & 42.3 & 31.0 & 5.8 & 16.1 & 7.3 & 28.7 & 24.3 & 30.3 & 8.0 & 7.9 & 6.2 & 20.2 \\
\addlinespace[3pt]
ZEBRA: reference      & 44.8 & \underline{43.2} & \underline{44.2} & \textbf{48.1} & 43.1 & \textbf{42.1} & \underline{47.7} & \underline{48.0} & \underline{46.7} & \textbf{32.2} & 19.8 & 16.5 & \underline{39.7} \\
ZEBRA: VLM partition  & \textbf{50.6} & \textbf{49.4} & \textbf{47.5} & 41.9 & \underline{43.4} & 37.6 & \textbf{56.4} & \textbf{54.8} & \textbf{55.0} & 23.7 & \underline{22.6} & \underline{20.5} & \textbf{42.0} \\
ZEBRA: cubic chunk    & 26.2 & 27.3 & 24.8 & \underline{44.4} & \textbf{43.6} & \underline{41.6} & 36.2 & 33.2 & 35.9 & \underline{27.5} & \textbf{33.5} & \textbf{26.6} & 33.4 \\
\bottomrule
\end{tabular}
\par\vspace{2pt}\raggedright\scriptsize \rev{Each column leaves 50, 100 or 150\,MB of device memory free beyond the residual budget of that pairing (Table~\ref{tab:devices}), so the map is given that much less. In every column \textbf{bold} is the best and \underline{underline} the second best.}
\end{table*}

\begin{table*}[t]
\centering
\begin{minipage}[t]{0.47\textwidth}
\centering
\caption{\rev{Payload loaded and ATE RMSE by map, mean over the devices and budgets of Table~\ref{tab:mission}.}}
\label{tab:cost}
\footnotesize
\setlength{\tabcolsep}{5pt}
\renewcommand{\arraystretch}{1.15}
\begin{tabular}{@{}lrrrr@{}}
\toprule
& \multicolumn{2}{c}{Hospital} & \multicolumn{2}{c}{Office} \\
\cmidrule(lr){2-3}\cmidrule(lr){4-5}
Condition & Load$\downarrow$ & RMSE$\downarrow$ & Load$\downarrow$ & RMSE$\downarrow$ \\
          & (MB) & (m) & (MB) & (m) \\
\midrule
Org: RTAB-Map         & 1924 & 5.84 & 1560 & 6.76 \\
Org: RTAB-Map, route  & 1959 & 5.62 & 1535 & 4.63 \\
\addlinespace[3pt]
ZEBRA: reference      & 2904 & 0.27 & 1882 & 1.82 \\
ZEBRA: VLM partition  & 2839 & 0.33 & 1425 & 1.57 \\
ZEBRA: cubic chunk    & 3279 & 0.57 & 1720 & 1.90 \\
\bottomrule
\end{tabular}
\par\vspace{2pt}\raggedright\scriptsize \rev{Load: payload read from the database over the run. RMSE: of the planar ATE over the localized frames, mean over the runs.}
\end{minipage}\hfill
\begin{minipage}[t]{0.50\textwidth}
\centering
\caption{\rev{Throughput of the concurrent model, tokens per second.}}
\label{tab:tps}
\fontsize{7.5}{9}\selectfont
\setlength{\tabcolsep}{2pt}
\renewcommand{\arraystretch}{1.35}
\begin{tabular}{@{}lrrccrr@{}}
\toprule
& \multicolumn{2}{c}{only VLM} & prev(RTAB-Map) & \multicolumn{3}{c}{Proposed (ZEBRA)} \\
\cmidrule(lr){2-3}\cmidrule(lr){4-4}\cmidrule(lr){5-7}
Pairing & Median & p10 & Outcome & Outcome & Median & p10 \\
\midrule
AGX Orin, hospital  & 2.09  & 2.08  & $\times$ & $\circ$ & 2.13  & 2.10  \\
AGX Orin, office    & 2.09  & 2.09  & $\times$ & $\circ$ & 2.12  & 2.10  \\
Orin Nano, hospital & 44.88 & 44.38 & $\times$ & $\circ$ & 44.61 & 44.31 \\
Orin Nano, office   & 28.20 & 27.70 & $\times$ & $\circ$ & 27.99 & 27.47 \\
\bottomrule
\end{tabular}
\par\vspace{2pt}\raggedright\scriptsize \rev{$\circ$: the mission completes. $\times$: stalled by an out-of-memory kill.}
\end{minipage}
\end{table*}

\begin{table*}[t]
\centering
\begin{minipage}[t]{0.48\textwidth}
\centering
\caption{Cubic chunk by side, mean over three pairings at the residual budget.}
\label{tab:side}
\small
\setlength{\tabcolsep}{5pt}
\begin{tabular}{@{}lrrrrr@{}}
\toprule
 & 2\,m & 4\,m & 6\,m & 8\,m & 10\,m \\
\midrule
Localized (\%)$\uparrow$          & 17.5 & 22.1 & 39.1 & 36.8 & 33.6 \\
Read (MB)$\downarrow$               &  794 & 1399 & 1941 & 2200 & 2412 \\
Reloads$\downarrow$                 &   32 &   48 &   82 &  104 &   93 \\
\bottomrule
\end{tabular}
\end{minipage}\hfill
\begin{minipage}[t]{0.48\textwidth}
\centering
\caption{Admission at the boundary against prefetch guided by the route, reference partition, mean over three pairings at the residual budget.}
\label{tab:route}
\small
\begin{tabular}{@{}lrr@{}}
\toprule
 & At the boundary & Guided by route \\
\midrule
Localized (\%)$\uparrow$              & 40.0 & 45.6 \\
Keyframes loaded (\#)$\downarrow$        & 572  & 560  \\
Keyframes unloaded (\#)$\downarrow$      & 251  & 231  \\
\bottomrule
\end{tabular}
\end{minipage}
\end{table*}

\section{EXPERIMENTS}
\rev{This section asks the two questions of the paper in turn, whether bounding the resident map in memory lets it share a device with a language model, and what the map should admit once that bound is enforced.
Section~\ref{subsec:setup} describes the setup, Sections~\ref{subsec:slam} and \ref{subsec:ai} report what each condition localizes and what the model beside it delivers, and Section~\ref{subsec:ablation} separates the side of the cubic chunk and the use of the route.}

\subsection{Experimental Setup}
\label{subsec:setup}

\subsubsection{Environments and maps}
\rev{We evaluate on two indoor environments built in NVIDIA Isaac Sim, a hospital and an office floor.
Each map is built offline from a mapping run that visits every room, and holds 1620 and 1041 keyframes and, with the point cloud decoded at decimation one, 8399 and 5263\,MB of resident payload (Table~\ref{tab:devices}).
Every run replays a query trajectory recorded by driving the mission below against the pre-built map with online mapping disabled, a wheeled robot carrying an RGB-D camera at 5\,Hz and moving at 0.6\,m/s.}

\subsubsection{Partitions}
\rev{Figures~\ref{fig:routes} and \ref{fig:partitions} show the routes and the three partitions.
The reference partition was drawn offline by Claude Fable (Anthropic) from a top-down view, with the long hall of the hospital cut at 8\,m and the open floor of the office at 10\,m so that no zone runs the length of the building, and used unchanged, 61 and 43 zones.
We call it a reference rather than a ground truth, since it encodes one reading of the layout. 
The second, the VLM partition, was drawn offline by a vision-language model, Gemma4 26B~\cite{ollama_gemma4}, with no access to the layout, from a top-down rendering of the occupancy grid and corrected once by the same model shown its own answer, 92 and 83 zones, finer because it splits large rooms at furniture boundaries, and its largest zone holds 172 and 164 keyframes against 271 and 159 for the reference partition.
The cubic chunk has no semantic content, a square cell of one fixed side admitted when the robot occupies it, its side set so that the chunks holding free space come within two of the zone count of the reference partition, 7\,m and 8.5\,m, so that the chunk does not lose to the reference partition merely by holding fewer and larger zones.}

\subsubsection{Devices and budgets}
\rev{Table~\ref{tab:devices} lists the two platforms, the model each runs, and the budget given to the map.
The models are served through Ollama~\cite{ollama_gemma4,ollama_qwen35,ollama_gemma3} and answer queries throughout every run, each query asking the model to describe a camera frame taken from the mission, or to describe the building in text where the model is text-only.
The model stays resident, as it would on a deployed robot, and the map is given only what it leaves free, the memory free while the model alone is answering queries, less the fixed part of the map process and a margin.
The Orin Nano with the hospital map has no feasible point under Gemma4 E2B, since the fixed part, 1.7\,GB, and the model, 4.2\,GB, leave the map no payload to hold, so that pairing runs the smallest model the board holds next to this map, Gemma3 1B, which is text-only.
Each pairing is run at the residual budget of Table~\ref{tab:devices} and, for Table~\ref{tab:mission}, with 50, 100 and 150\,MB of device memory left free beyond it and taken from the map, and the payload of each map is 2.5--10.5 times its residual budget and up to 15 times the residual less 150\,MB, so no configuration holds more than a fraction of the map.}

\subsubsection{Conditions}
\rev{Every condition runs under the same memory budget.
The two conditions written Org are the load and unload algorithm of RTAB-Map itself, which acquires the graph neighbors of the keyframe the current frame matched, in the order the links give, altered only so that it holds the budget strictly, which that system does not do in localization mode, and carrying no zone partition and no eviction policy of our own.
The route variant is the same mechanism given the route of the mission, so it retrieves along the path it plans and holds the budget in that form.
Every condition starts from the same true pose, and what differs is what each turns it into.}

\subsubsection{Scenario}
\rev{The robot starts in the lobby, travels to a designated room, stays there for 55\,s while the model keeps answering queries, and returns, and it is run to three rooms on each floor, the routes R1 to R3 of Fig.~\ref{fig:routes}.
We report the localization rate, the percentage of query frames for which a pose was produced, and the absolute trajectory error (ATE), the root mean square of the planar error against the simulator's ground-truth pose over the localized frames.}

\subsection{Effect of the Memory Budget on Localization}
\label{subsec:slam}

\textbf{\rev{Mission comparison.}}
\rev{The budget holds, since no run exceeded it in any recorded cycle and none was killed, whereas without it the operating system ends one of the two processes on every pairing.
How hard it presses differs by pairing and by route, since zones are retired in every run of the AGX Orin with the hospital map and of the Orin Nano with the office map, while on the other two pairings the pressure falls on the route to the farthest room.
On the Orin Nano with the office map the budget is below the largest zone on the route, so the fit property of Section~\ref{subsec:properties} does not hold for the reference partition, which appears to be why the semantic conditions do not lead on that pairing.}

\rev{Under that budget, what is resident decides whether the mission succeeds.
Over the twelve cells of Table~\ref{tab:mission} the original retrieval localizes about 20\,\% of the query frames, the cubic chunk 33.4\,\% and the two semantic partitions 39.7 and 42.0\,\%, 6--9 points ahead of the chunk at a matched zone count, and the ATE separates retrieval from the zone conditions by an order of magnitude on the hospital map and by a factor of three on the office map (Table~\ref{tab:cost}).}

\rev{The retrieval produces no pose in the target room in 58 of the 72 runs of the two original conditions and under 4\,\% of the frames spent there in six more, so the rate comes from the two corridor legs, and it holds the room only in the remaining eight, 44--87\,\%.
On the leg out of the first route the twenty-four runs of the two original conditions fall into two groups with nothing between them and with no ordering by the budget, ten that hold the match from the first cycle and localize 24--40\,\% of the leg, and fourteen that lose it and localize 1--7\,\%.
The retrieval reads the graph neighbors of the keyframe it last matched, so a cycle that fails to match leaves the same neighborhood resident, the next cycle fails for the same reason, and a lost run recovers only on the leg back where the keyframes read at the start are still resident.
Zone admission has no such loop, because what is read is decided by the pose and the zone it falls in.}

\rev{Table~\ref{tab:mission} averages the three routes run one at a time, and we also ran them as one continuous mission, R1 to R3 with a return through the lobby between them, on all four pairings at the residual budget, 685 and 643 seconds of replay.
The ordering of Table~\ref{tab:mission} holds there as well.}

\rev{Between the two semantic partitions, the one drawn by the vision-language model leads on the hospital map and loads less to do so, because its largest zone holds fewer keyframes than the reference partition does on that map and an admission therefore fits whole more often, while on the office map the largest zone of the two is nearly the same size, 164 against 159 keyframes, and they are within seven points on the AGX Orin, where every admission fits whole.
What this says about the partition is that a room as a person would name it is not always the right zone, since a large room costs more to admit than the robot needs at once, and dividing it further along a boundary the space itself gives, furniture in the case of the automatic partition, keeps the zone aligned with visibility while bringing the admission inside the budget.
On the hospital map every zone condition gains 3--11 points on the Orin Nano over the AGX Orin, the same map under a smaller model that leaves the map a larger budget (Table~\ref{tab:devices}), and the cubic chunk is last on that map and loads the most of the three.}

\subsection{Effect of the Memory Budget on the Concurrent AI Workload}
\label{subsec:ai}
\rev{The memory the map is given is taken from the workload beside it, and without the bound that workload is lost (Table~\ref{tab:tps}).
The prev column is the system as shipped, and the ZEBRA columns are the per-run median and tenth percentile over the runs of every condition.
Unbounded, the operating system ends one of the two processes on every pairing, the model on three of them and the map on the Orin Nano with the hospital map, after which the model answers to the end and the mission localizes nothing.
On both Orin Nano pairings the throughput fell while the map was still resident, to a tenth percentile of 43.8 over the calls completed in that window against the 44.88 the model reaches with no map resident on the hospital map, and 25.0 against 28.20 on the office map before the model itself was ended.
On the AGX Orin the model is killed before enough calls complete for a decline to be visible, so what that board shows is not a slow degradation but an outright kill, and the memory bandwidth of the two boards differs by a factor of two~\cite{jetson_agx_orin,jetson_orin_nano}.
Under ZEBRA every pairing completes with both processes alive and the model gives up at most 0.8 percent of the rate it reaches with no map resident.
The tenth percentile stays within 0.03 of the median on the AGX Orin and within 0.6 on the Orin Nano.}

\subsection{Ablation on Chunk Side and Route}

\label{subsec:ablation}

\textbf{\rev{The side of the chunk.}}
\rev{No single side serves every map.
The cost rises with the side because a larger chunk is a larger admission and a larger re-read on the return leg, and beyond 6\,m a side reads more and localizes less, while the smallest side reads the least and localizes the least, since a small chunk multiplies boundary crossings and a robot that has lost the map admits nothing.
The best side is a different one on each device and map, so it has to be chosen for each of them and is not known before the run.}

\textbf{\rev{Route-conditioned admission.}}
\rev{Every zone condition above was run with route-conditioned admission, and here the route is switched off.
Table~\ref{tab:route} compares reading a zone when its boundary is crossed against reading the next zone on the route while the robot is still in the current one, on the reference partition and the same three pairings.
Against admission at the boundary with no route at all, reading ahead raises the localization rate by 5.6 points, because a zone read before its boundary is resident when the boundary is crossed, and on average it admits and releases less, because admission at the boundary follows a pose estimate that drifts when localization is lost.}

\section{CONCLUSION}
We propose a map-management layer that bounds the resident map in bytes rather than in a count of keyframes, and that admits and releases keyframes a zone at a time.
The unit of accounting is the byte, the unit of admission is the semantic zone, and every cycle measures the resident total and makes room before any data is read, so the budget is never overshot.

\rev{Accounting in memory itself is what removes the out-of-memory failure, since with the budget enforced no run exceeded it, none was killed, and the model gave up at most 0.8 percent of the rate it reaches with no map resident, whereas without it the operating system ends the model or the map on every pairing.
Under that budget the semantic partitions localize more of the query frames than the geometric one in these environments, 39.7 and 42.0 percent of a mission to a designated room, ahead of the 33.4 of a fixed grid of matched zone count and of about 20 for keyframe-level retrieval, at a lower ATE.

Future work follows two directions.
The first is composition with scene-graph systems that already divide a building into rooms, which could draw the partition online while the map is still being built, so that the layer could apply where the robot maps and localizes at once and not only against a map built beforehand.
The second is composition with work that puts semantic space to other uses. What this paper contributes is the memory side of that division, and a robot that already reasons over rooms and corridors to decide where to go or what to answer can carry the same partition into residency, which is where we expect the combination to gain the most.}

\section*{ACKNOWLEDGMENT}
\noindent\textbf{Use of generative AI.}
Claude (Anthropic) was used to translate and revise the authors' drafts and for grammar and language editing, and to write the scripts that render the figures and summarize the run logs.
It also drew the reference partition of Section~\ref{subsec:setup} from a top-down view of the map, which is an input the evaluation compares and not a source of any reported value.
It produced no measurement, all experiments were run by the authors on the platforms of Section~\ref{subsec:setup}, every reported value was verified against the run logs, and the figures render measured data.
The vision-language model that draws the second partition is part of the method under study, not an authoring tool.


{\footnotesize
\bibliography{references}
}

\end{document}